\documentclass[sigconf]{acmart}

\copyrightyear{2019} 
\acmYear{2019} 
\setcopyright{acmcopyright}
\acmConference[KDD '19]{The 25th ACM SIGKDD Conference on Knowledge Discovery and Data Mining}{August 4--8, 2019}{Anchorage, AK, USA}
\acmBooktitle{The 25th ACM SIGKDD Conference on Knowledge Discovery and Data Mining (KDD '19), August 4--8, 2019, Anchorage, AK, USA}
\acmPrice{15.00}
\acmDOI{10.1145/3292500.3330871}
\acmISBN{978-1-4503-6201-6/19/08}

\usepackage{booktabs} 

\usepackage{graphicx}
\usepackage{amssymb}
\usepackage{amsthm}
\usepackage{adjustbox}
\usepackage[font=small]{caption}
\usepackage{multirow} 

\usepackage{algorithmic}
\usepackage{algorithm}
\usepackage{textcase} 

\begin{document}

\title[Deep Anomaly Detection with Deviation Networks]{Deep Anomaly Detection with Deviation Networks}

\author{Guansong Pang}
\authornote{Guansong Pang is the corresponding author.}
\affiliation{%
  \department{Australian Institute for Machine Learning}
  \institution{The University of Adelaide}
  \city{Adelaide}
  \country{Australia}
  \postcode{5005}
}
\email{guansong.pang@adelaide.edu.au}

\author{Chunhua Shen}
\affiliation{%
  \department{Australian Institute for Machine Learning}
  \institution{The University of Adelaide}
  \city{Adelaide}
  \country{Australia}
  \postcode{5005}
}
\email{chunhua.shen@adelaide.edu.au}

\author{Anton van den Hengel}
\affiliation{%
  \department{Australian Institute for Machine Learning}
  \institution{The University of Adelaide}
  \city{Adelaide}
  \country{Australia}
  \postcode{5005}
}
\email{anton.vandenhengel@adelaide.edu.au}

\begin{abstract}

Although deep learning has been applied to successfully address many data mining problems, relatively limited work has been done on deep learning for anomaly detection. Existing deep anomaly detection methods, which focus on learning new feature representations to enable downstream anomaly detection methods, perform indirect optimization of anomaly scores, leading to data-inefficient learning and suboptimal anomaly scoring. Also, they are typically designed as unsupervised learning due to the lack of large-scale labeled anomaly data. As a result, they are difficult to leverage prior knowledge (e.g., a few labeled anomalies) when such information is available as in many real-world anomaly detection applications. 

This paper introduces a novel anomaly detection framework and its instantiation to address these problems. Instead of representation learning, our method fulfills an end-to-end learning of anomaly scores by a neural deviation learning, in which we leverage a few  (e.g., multiple to dozens) labeled anomalies and a prior probability to enforce statistically significant deviations of the anomaly scores of anomalies from that of normal data objects in the upper tail. Extensive results show that our method can be trained substantially more data-efficiently and achieves significantly better anomaly scoring than state-of-the-art competing methods.


\end{abstract}



\maketitle

\section{Introduction}

Anomalies are referred to as data objects that deviate significantly from the majority of data objects. Anomaly detection is the task of identifying these anomalies, which has important applications in broad domains, e.g., to detect network attacks in cybersecurity, to detect fraudulent transactions in finance, and to detect diseases in healthcare. Numerous anomaly detection methods have been introduced, but they often fail to detect anomalies in data with high dimensionality and/or intricate relations due to the curse of dimensionality and highly non-linear feature relations \cite{ruff2018deepsvdd,pang2018repen}. 

In recent years, deep learning \cite{lecun2015deep} has gained exceptional successes in discovering such intricate relations in high-dimensional data. However, deep learning for anomaly detection has been insufficiently explored due to the following two major challenges: (i) it is very difficult to obtain large-scale labeled data to train anomaly detectors due to the prohibitive cost of collecting such data in many anomaly detection application domains; and (ii) anomalies often demonstrate different anomalous behaviors, and as a result, they are dissimilar to each other, which poses significant challenges to widely-used optimization objectives that generally assume the data objects within each class are similar to each other. 

\begin{figure}[h!]
  \centering
    \includegraphics[width=0.485\textwidth]{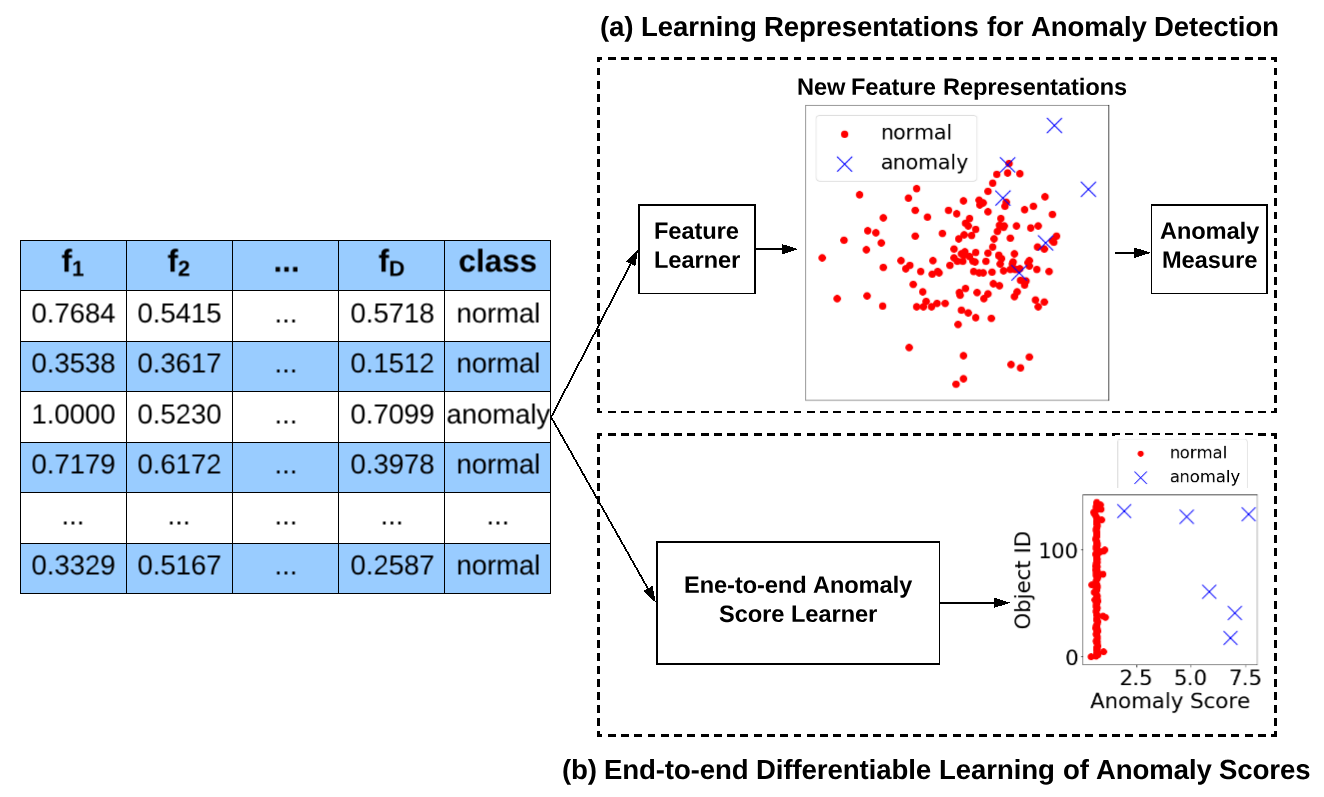}
  \caption{(a) Learning Features for Subsequent Anomaly Measures vs. (b) Direct Learning of Anomaly Scores}
  \label{fig:example}
\end{figure}

Existing deep anomaly detection\footnote{Deep anomaly detection refers to any methods that exploit deep learning techniques to learn feature representations or anomaly scores for anomaly detection.} methods \cite{hawkins2002autoencoder,zhou2017autoencoder,chen2017autoencoder,schlegl2017gan,zenati2018gan,pang2018repen,ruff2018deepsvdd} address these two challenges by using unsupervised deep learning to model the normal class in a two-step approach (i.e., the pipeline (a) in Figure \ref{fig:example}): they first learn to represent data with new representations, e.g., intermediate representations in autoencoders \cite{hawkins2002autoencoder,zhou2017autoencoder,chen2017autoencoder}, latent spaces in generative adversarial networks (GANs) \cite{schlegl2017gan,zenati2018gan}, or distance metric spaces in \cite{pang2018repen,ruff2018deepsvdd}; and then they use the learned representations to define anomaly scores using reconstruction errors \cite{hawkins2002autoencoder,zhou2017autoencoder,chen2017autoencoder,schlegl2017gan,zenati2018gan} or distance-based measures \cite{pang2018repen,ruff2018deepsvdd}. However, in most of these methods \cite{hawkins2002autoencoder,zhou2017autoencoder,chen2017autoencoder,schlegl2017gan,zenati2018gan}, the representation learning is separate from anomaly detection methods, so it may yield representations that are suboptimal or even irrelevant w.r.t. specific anomaly detection methods. The very recent efforts \cite{pang2018repen,ruff2018deepsvdd} address this problem by incorporating traditional anomaly scoring measures into the feature learning objective, which substantially improves the expressiveness of the feature representations. However, they still focus on optimizing the representations, which is an indirect optimization of anomaly scoring. This can lead to inefficient use of training data and low-quality anomaly scoring. 

Also, they are mainly focused on unsupervised learning, which may lead to a common problem of unsupervised anomaly detection that many of the anomalies they identify are data noises or uninteresting data objects due to the lack of prior knowledge of the anomalies of interest \cite{aggarwal2017supervised,pang2018repen,siddiqui2018kdd}. A potential solution to this problem is to leverage a limited number of labeled anomalies as the prior knowledge to learn anomaly-informed models, since such prior knowledge is often available in many real-world anomaly detection applications. Those labeled anomalies may originally come from a deployed detection system, e.g., a few successfully detected network intrusion records, or they may be from users, such as a small number of fraudulent credit card transactions that are reported by clients and confirmed by the banks.


In this paper, we introduce a novel anomaly detection framework to fill these gaps by leveraging a few labeled anomalies to fulfill an end-to-end differentiable learning of anomaly scores. That is, as shown in the pipeline (b) in Figure \ref{fig:example}, with the original data as inputs, we directly learn and output the anomaly scores rather than the feature representations. Specifically, as shown in Figure \ref{fig:framework}, given a training data object, the proposed framework first uses a neural anomaly score learner to assign it an anomaly score, and then defines the mean of the anomaly scores of some normal data objects based on a prior probability to serve as a reference score for guiding the subsequent anomaly score learning. Lastly, the framework defines a loss function, called deviation loss, to enforce statistically significant deviations of the anomaly scores of anomalies from that of normal data objects in the upper tail.

We further instantiate the framework into a method called deviation networks (DevNet). DevNet leverages multiple to dozens of labeled anomalies, accounting for only 0.005\%-1\% of all training data objects and 0.08\%-6\% of all anomalies per data set, and a Gaussian prior to perform a direct optimization of anomaly scores using a Z-Score-based deviation loss. By doing so, DevNet can not only achieve very data-efficient learning of the anomaly scores but also accommodate anomalies with different anomalous behaviors. Additionally, in contrast to most methods that produce hardly interpretable anomaly scores \cite{kriegel2011interpreting}, the Z-Score-based deviation loss also allows DevNet to produce easily interpretable anomaly scores.

Accordingly, this paper makes the following major contributions.
\begin{itemize}
    \item We introduce a novel framework to learn anomaly scores in an end-to-end fashion. In contrast to the current indirect optimization approach, our framework fulfills a direct optimization of anomaly scores. As far as we know, this is the first framework for leveraging limited labeled anomaly data to achieve end-to-end anomaly score learning.
    \item A novel anomaly detection method, namely deviation networks (DevNet\footnote{Our code is made available at https://sites.google.com/site/gspangsite/sourcecode.}), is instantiated from the framework. DevNet synthesizes neural networks, Gaussian prior and Z-Score-based deviation loss to perform data-efficient and effective learning of the anomaly scores, resulting in well optimized and easily interpretable anomaly scores.
\end{itemize}
Extensive empirical results on nine large and/or high-dimensional real-world data sets show that (i) DevNet significantly outperforms four state-of-the-art competing methods in terms of both the Area Under Receiver Operating Characteristic Curve (AUC-ROC) and Precision-Recall curve (AUC-PR), with 3\%-29\% average AUC-ROC improvement and 21\%-309\% average AUC-PR improvement; and (ii) DevNet obtains a substantially better data efficiency than the competing methods, e.g., it can use 75\%-88\% less labeled anomalies to achieve the accuracy that is comparably good to, or substantially better than, the best contenders. 

\section{Related Work}
\subsection{Traditional Anomaly Detection}

Most traditional anomaly detection approaches, e.g., distance-based approach and density-based approach, are ineffective in handling irrelevant features or non-linear separable classes due to the curse of dimensionality and the deficiency in capturing the non-linear relations. Recently ensemble methods (e.g., iForest \cite{liu2012iforest} and many others \cite{keller2012hics,pang2018sparse}) showed some large improvement over these approaches by working on selected feature subspaces, but how to efficiently and effectively identify the relevant subspaces and model the intricate relations is still an open problem in anomaly detection. 

\subsection{Deep Anomaly Detection}

Current popular deep anomaly detection methods are unsupervised approach, including autoencoder-based methods and GANs-based methods. Autoencoder-based methods \cite{hawkins2002autoencoder,chen2017autoencoder,zhou2017autoencoder} use a bottleneck network architecture to learn a low-dimensional representation space, and then use the learned representations to define reconstruction errors as anomaly scores. GANs-based methods \cite{schlegl2017gan,zenati2018gan} also use the reconstruction error as anomaly score, but they leverage two competing networks, a generator and a discriminator, to adversarially learn a latent space of the training data and use this latent space to compute the reconstruction errors. These deep methods can capture more complex feature interactions than traditional shallow methods such as random projection \cite{li2006srp}, but they learn the representations separately from the subsequent anomaly detection, leading to suboptimal or unstable detection performance \cite{pang2018repen,ruff2018deepsvdd}. To address this issue, very recent work \cite{pang2018repen,ruff2018deepsvdd} focuses on unifying the representation learning and anomaly detection. The REPEN method \cite{pang2018repen} exploits triplet networks to integrate the representation learning with distance-based detectors, while deep Support Vector Data Description (SVDD) \cite{ruff2018deepsvdd} aims to learn representations for the one-classifier, SVDD \cite{tax2004svdd}. Both REPEN and deep SVDD achieve substantial improvement over the previous methods. However, their optimization objective still focuses on feature representations, so they optimize the anomaly scoring in an indirect manner. DevNet is fundamentally different from these methods in that DevNet performs a direct differentiable learning of the anomaly scores in an end-to-end fashion.


\subsection{Anomaly Detection with Limited Data}
Only a few studies have been done on performing anomaly detection with a few labeled anomalies. In \cite{mcglohon2009snare,tamersoy2014guilt}, a small set of labeled anomalies is incorporated into a belief propagation process to achieve more reliable anomaly scoring, but they are only applicable to graph data. In \cite{pang2018repen}, REPEN leverages a few labeled anomalies to learn more application-relevant feature representations, resulting in over 30\% accuracy improvement compared to its fully unsupervised version.

This research line is relevant to few-shot classification \cite{fei2006oneshot,snell2017protonet} and PU learning (i.e., learning from positive and unlabeled examples) \cite{li2003pul,elkan2008pul,sansone2018pul}. Few-shot classification is relevant because it also aims to leverage a few labeled examples to identify incoming objects of the same class. However, they are very different because (i) in few-shot classification, we have a large number of labeled data of the seen classes during training, but we do not know any class information of the training data in anomaly detection; and (ii) few-shot classification implicitly assumes that the few labeled objects and incoming objects of each of the unseen classes share the same manifold, whereas the few labeled anomalies and the unseen anomalies may be from very different manifolds. The second difference is also the key difference between our task and PU learning, because PU learning also has the same assumption as few-shot classification since they are both focused on classification. Also, most PU learning techniques typically require a substantially large percentage of positive examples to work well, e.g., 45\% in \cite{li2003pul}, 20\%-50\% in \cite{elkan2008pul} and 20\% in \cite{sansone2018pul}, which is often not practical or too costly to collect that much anomaly data in many anomaly detection applications. Therefore, both few-shot and PU learning techniques are significantly challenged by the studied problem.




\section{End-to-end Anomaly Score Learning}

\subsection{Problem Statement}

Instead of taking the current two-step approach that first learns new representations and then applies some anomaly measures to the new representations to compute anomaly scores, we aim to leverage a small number of labeled anomalies to directly learn the anomaly scores. Specifically, given a set of $N+K$ training data objects $\mathcal{X}=\{ \mathbf{x}_{1}, \mathbf{x}_{2}, \cdots, \mathbf{x}_{N}, \mathbf{x}_{N+1}, \mathbf{x}_{N+2}, \cdots, \mathbf{x}_{N+K} \}$ with $\mathbf{x}_{i} \in \mathbb{R}^{D}$, in which $\mathcal{U}=\{ \mathbf{x}_{1}, \mathbf{x}_{2}, \cdots, \mathbf{x}_{N}\}$ is unlabeled data and $\mathcal{K}=\{\mathbf{x}_{N+1}, \mathbf{x}_{N+2}, \cdots, \mathbf{x}_{N+K} \}$ with $K\ll N$ is a very small set of labeled anomalies that provide some prior knowledge of anomalies, our goal is to learn an anomaly scoring function $\phi: \mathcal{X} \mapsto \mathbb{R}$ that assigns anomaly scores to data objects in a way that we have $\phi(\mathbf{x}_{i}) > \phi(\mathbf{x}_{j})$ if $\mathbf{x}_{i}$ is an anomaly and $\mathbf{x}_{j}$ is a normal data object. 

\subsection{The Proposed Framework}

To solve this problem, we introduce a novel framework that synthesizes neural networks, a prior probability distribution of anomaly scores, and a new loss function to train a deep anomaly detector in an end-to-end fashion, with an objective to assign statistically significantly larger anomaly scores to anomalies than normal objects. The resulting model is expected to yield more optimized anomaly scores and be more data-efficient than the two-step approach. 

\subsubsection{The Procedure of the Framework}
\begin{figure}[h!]
  \centering
    \includegraphics[width=0.46\textwidth]{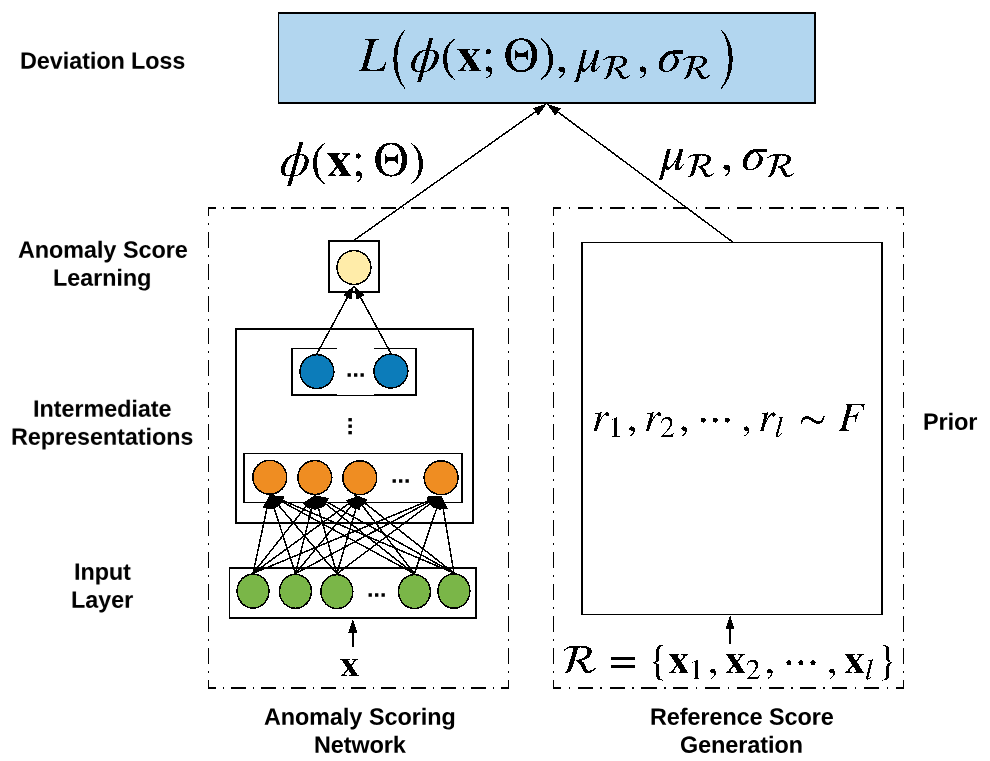}
  \caption{The Proposed Framework. $\phi(\mathbf{x}; \Theta)$ is an anomaly score learner with the parameters $\Theta$. $\mu_{\mathcal{R}}$ is the mean of the anomaly scores of some normal objects, which is determined by a prior $F$. $\sigma_{\mathcal{R}}$ is a standard deviation associated with $\mu_{\mathcal{R}}$. The loss $L\big(\phi(\mathbf{x}; \Theta), \mu_{\mathcal{R}}, \sigma_{\mathcal{R}}\big)$ is defined to guarantee that the anomaly scores of anomalies statistically significantly deviate from $\mu_{\mathcal{R}}$ in the upper tail while enforce normal objects have anomaly scores as close as possible to $\mu_{\mathcal{R}}$. }
  \label{fig:framework}
\end{figure}

As shown in Figure \ref{fig:framework}, our framework consists of three major modules: 
\begin{enumerate}
    \item We first use an \textit{anomaly scoring network}, i.e., a function $\phi$, to yield a scalar anomaly score for every given input $\mathbf{x}$.
    \item To guide the learning of anomaly scores, we then use a \textit{reference score generator} to generate another scalar score termed as \textit{reference score}, which is defined as the mean of the anomaly scores $\{r_1, r_2, \cdots, r_l\}$ for a set of $l$ randomly selected normal objects, denoted as $\mu_{\mathcal{R}}$. The reference score $\mu_{\mathcal{R}}$ may be either learned from a model or determined by a prior probability $F$. The latter one is chosen so as to efficiently generate $\mu_{\mathcal{R}}$ and obtain interpretable anomaly scores. 
    \item Lastly $\phi(\mathbf{x})$, $\mu_{\mathcal{R}}$ and its associated standard deviation $\sigma_{\mathcal{R}}$ are input to the \textit{deviation loss} function $L$ to guide the optimization, in which we aim to optimize the anomaly scores so that the scores of anomalies statistically significantly deviate from $\mu_{\mathcal{R}}$ in the upper tail while at the same time having the scores of normal objects as close as possible to $\mu_{\mathcal{R}}$.
\end{enumerate}
 
One problem here is how to effectively obtain a sufficient number of normal objects to train our model, since we only have a few labeled anomalies in $\mathcal{K}$ but do not know the class label of the objects in $\mathcal{U}$. We will discuss how to address this problem in Section \ref{sec:loss}.

\subsubsection{How Does The Proposed Framework Address the Aforementioned Two Main Challenges of Deep Anomaly Detection?}

The deviation loss-based optimization in our framework forces the normal objects cluster around $F$ in terms of their anomaly scores but pushes anomalies statistically far away $F$, which well optimizes the anomaly scores and also empowers the intermediate representation learning to discriminate normal objects from the rare anomalies with different anomalous behaviors. In other words, our deep anomaly detector leverages a few labeled anomalies and the prior of anomaly scores to learn a high-level abstraction of normal behaviors, enabling it to assign a large anomaly score to an object as long as the object's behaviors significantly deviate from the learned abstraction of being normal. This offers an effective detection of dissimilar anomalies, e.g., anomalies due to different reasons or previously unknown anomalies; and in turn the optimization also requires substantially less labeled anomalies to train the detector.

\section{Deviation Networks}

The proposed framework is instantiated into a method called Deviation Networks (DevNet), which defines a Gaussian prior and a Z-Score-based deviation loss to enable the direct optimization of anomaly scores with an end-to-end neural anomaly score learner.

\subsection{End-to-end Anomaly Scoring Network}
Let $\mathcal{Q} \in \mathbb{R}^{M}$ be an intermediate representation space, an anomaly scoring network $\phi(\cdot; \Theta):\mathcal{X} \mapsto \mathbb{R}$ can be defined as a combination of a feature representation learner $\psi(\cdot; \Theta_{t}): \mathcal{X} \mapsto \mathcal{Q}$ and an anomaly scoring function $\eta(\cdot; \Theta_{s}): \mathcal{Q} \mapsto \mathbb{R}$, in which $\Theta=\{\Theta_{t}, \Theta_{s}\}$. 

Specifically, $\psi(\cdot; \Theta_{t})$ is a neural \textit{feature learner} with $H \in \mathbb{N}$ hidden layers and their weight matrices $\Theta_{t}=\{\mathbf{W}^{1}, \mathbf{W}^{2}, \cdots, \mathbf{W}^{H}\}$, which can be represented as
\begin{equation}
    \mathbf{q} = \psi(\mathbf{x}; \Theta_{t}),
\end{equation}
where $\mathbf{x} \in \mathcal{X}$ and $\mathbf{q} \in \mathcal{Q}$. Different hidden network structures can be used here based on the type of data inputs, such as multilayer perceptron networks for multidimensional data, convolutional networks for image data, or recurrent networks for sequence data. 

$\eta(\cdot, \Theta_{s}): \mathcal{Q} \mapsto \mathbb{R}$ is defined as an \textit{anomaly score learner} which uses a single linear neural unit in the output layer to compute the anomaly scores based on the intermediate representations:

\begin{equation}
    \eta(\mathbf{q};\Theta_{s}) = \sum_{i=1}^{M}w^{o}_{i} q_{i} + w^{o}_{M+1},
\end{equation}
where $\mathbf{q} \in \mathcal{Q}$ and $\Theta_{s} = \{\mathbf{w}^{o}\} $ ($w^{o}_{M+1}$ is the bias term).

Thus, $\phi(\cdot; \Theta)$ can be formally represented as 
\begin{equation}
    \phi(\mathbf{x};\Theta) = \eta(\psi(\mathbf{x};\Theta_{t});\Theta_s),
\end{equation}
which directly maps data inputs to scalar anomaly scores and can be trained in an end-to-end fashion.

\subsection{Gaussian Prior-based Reference Scores}\label{sec:reference}

Having obtained the anomaly scores using $\phi(\mathbf{x};\Theta)$, a \textit{reference score} $\mu_{\mathcal{R}} \in \mathbb{R}$, which is defined as the mean of the anomaly scores of a set of some randomly selected normal objects $\mathcal{R}$, is fed into the network output to guide the optimization. There are two main ways to generate $\mu_{\mathcal{R}}$: data-driven and prior-driven approaches. Data-driven methods involve a model to learn $\mu_{\mathcal{R}}$ based on $\mathcal{X}$, while prior-driven methods generate $\mu_{\mathcal{R}}$ from a chosen prior probability $F$. The prior-based approach is chosen here because (i) the chosen prior allows us to achieve good interpretability of the predicted anomaly scores and (ii) it can generate $\mu_{\mathcal{R}}$ constantly, which is substantially more efficient than the data-driven approach.

The specification of the prior is the main challenge of the prior-based approach. Fortunately, extensive results in \cite{kriegel2011interpreting} show that Gaussian distribution fits the anomaly scores very well in a range of data sets. This may be due to that the most general distribution for fitting values derived from Gaussian or non-Gaussian variables is the Gaussian distribution according to the central limit theorem. Motivated by this, we define a Gaussian prior-based reference score:
\begin{align}
    r_1, r_2, \cdots, r_l & \sim \mathcal{N}(\mu, \sigma^{2}), \\
    \mu_{\mathcal{R}} & = \frac{1}{l}\sum_{i=1}^{l}r_i,
\end{align}
where each $r_i$ is drawn from $\mathcal{N}(\mu, \sigma^{2})$ and represents an anomaly score of a random normal data object. We found empirically that DevNet was not sensitive to the choices of $\mu$ and $\sigma$ as long as $\sigma$ was not too large. We set $\mu=0$ and $\sigma=1$ in our experiments, which help DevNet to achieve stable detection performance on different data sets. DevNet is also not sensitive to $l$ when $l$ is sufficiently large due to the central limit theorem. $l=5000$ is used here.

\subsection{Z-Score-based Deviation Loss}\label{sec:loss}

A deviation loss is then defined to optimize the anomaly scoring network, with the deviation specified as a Z-Score
\begin{equation}
    \mathit{dev}(\mathbf{x}) = \frac{\phi(\mathbf{x};\Theta) - \mu_{\mathcal{R}}}{\sigma_{\mathcal{R}}},
\end{equation}
where $\sigma_{\mathcal{R}}$ is the standard deviation of the prior-based anomaly score set, $\{r_1, r_2, \cdots, r_l\}$. The deviation can then be plugged into the contrastive loss \cite{hadsell2006contrastloss} to specify our deviation loss as follows
\begin{equation}\label{eqn:loss}
    L\big( \phi(\mathbf{x};\Theta), \mu_{\mathcal{R}}, \sigma_{\mathcal{R}} \big) = (1-y)|\mathit{dev}(\mathbf{x})| + y \max\big(0, a - \mathit{dev}(\mathbf{x})\big),
\end{equation}
where $y=1$ if $\mathbf{x}$ is an anomaly and $y=0$ if $\mathbf{x}$ is a normal object, and $a$ is equivalent to a Z-Score confidence interval parameter. This loss enables DevNet to push the anomaly scores of normal objects as close as possible to $\mu_{\mathcal{R}}$ while enforce a deviation of at least $a$ between $\mu_{\mathcal{R}}$ and the anomaly scores of anomalies. Note that if $\mathbf{x}$ is an anomaly and it has a negative $\mathit{dev}(\mathbf{x})$, the loss is particularly large, which encourages large \textit{positive deviations} for all anomalies. Therefore, the deviation loss is equivalent to enforcing a statistically significant deviation of the anomaly score of all anomalies from that of normal objects in the upper tail. We use $a=5$ to achieve a very high significance level (i.e., 5.73303e-07) for all labeled anomalies. 

Similar to the contrastive loss, the deviation loss is monotonically increasing in $|\mathit{dev}(\mathbf{x})|$ and is monotonically deceasing in $\max\big(0, a - \mathit{dev}(\mathbf{x})\big)$, so it is convex w.r.t. both cases. However, they are also very different, because the contrastive loss uses pairs of intra-class/inter-class data objects as training samples to learn a similarity metric, whereas our deviation loss is built upon the deviation function and dedicated to the direct learning of anomaly scores.

One problem for using Eqn. (\ref{eqn:loss}) is that we do not have the labeled normal objects. We address this problem by simply treating the unlabeled training data objects in $\mathcal{U}$ as normal objects. Our empirical results showed that DevNet and also its competing deep methods performed very well by using this simple strategy, even when there was a large \textit{anomaly contamination level} (i.e., the proportion of anomalies in the unlabeled training data set $\mathcal{U}$). This may be because anomalies are rare data objects and their impacts become very limited on the stochastic gradient descent-based optimization in these deep detectors. Therefore, this training strategy is used by DevNet and its competing deep methods throughout our experiments. This can be seen as training the model with noisy data sets. We will evaluate the impact of different noise levels on the detection performance in Sections \ref{exp:contamination}.

\subsection{The DevNet Algorithm}\label{sec:algo}

Algorithm \ref{alg:devnet} presents the procedure of DevNet. After a random weight initialization in Step 1, DevNet performs stochastic gradient descent-based optimization to learn the weights in $\Theta$ in Steps 2-10. Particularly, Step 4 first samples a mini-batch $\mathcal{B}$ of size $b$ using stratified random sampling, followed by sampling the anomaly scores of $l$ normal objects from the prior $\mathcal{N}(\mu, \sigma^2)$ in Step 5. After obtaining $\mu_{\mathcal{R}}$ and $\sigma_{\mathcal{R}}$ in Step 6, Step 7 performs the forward propagation of the anomaly scoring network and computes the loss. Step 8 then performs gradient descent steps w.r.t. the parameters in $\Theta$. We finally obtain the optimized scoring network $\phi$.

\renewcommand{\algorithmicrequire}{\textbf{Input:}}
\renewcommand{\algorithmicensure}{\textbf{Output:}}
\begin{algorithm}
\small 
\caption{\textit{Training DevNet}}
\begin{algorithmic}[1]
\label{alg:devnet}
\REQUIRE $\mathcal{X} \in \mathbb{R}^{D}$ - training data objects, i.e., $\mathcal{X}=\mathcal{U} \cup \mathcal{K}$ and $\emptyset=\mathcal{U} \cap \mathcal{K}$ 
\ENSURE $\phi: \mathcal{X} \mapsto \mathbb{R}$ - an anomaly scoring network
\STATE Randomly initialize $\Theta$
\FOR{ $i = 1$ to $\mathit{n\_epochs}$}
    \FOR{ $j = 1$ to $\mathit{n\_batches}$}
        \STATE $\mathcal{B} \leftarrow$ Randomly sample $\mathit{b}$ data objects with a half of objects from $\mathcal{K}$ and another half from $\mathcal{U}$
        \STATE Randomly sample $l$ anomaly scores from $\mathcal{N}(\mu, \sigma^2)$
        \STATE Compute $\mu_{\mathcal{R}}$ and $\sigma_{\mathcal{R}}$ of the $l$ anomaly scores: $\{r_1, r_2, \cdots, r_l\}$
        \STATE $\mathit{loss} \leftarrow \frac{1}{b}\sum_{\mathbf{x} \in \mathcal{B}}L\big(\phi(\mathbf{x};\Theta), \mu_{\mathcal{R}}, \sigma_{\mathcal{R}} \big)$    
        \STATE Perform a gradient descent step w.r.t. the parameters in $\Theta$
    \ENDFOR
\ENDFOR
\RETURN $\phi$
\end{algorithmic}
\end{algorithm}

The core computation of training DevNet is the forward and backward propagation of the anomaly scoring network $\phi$, so the time complexity of DevNet depends on the network architecture used. For example, for multilayer perceptron networks, both the forward and backward propagation have the same complexity of $O(Dh_1+h_1h_2+\cdots+h_H*1)$, where $h_i$ is the number of neural units in the $i$-th hidden layer, so DevNet has an overall time complexity of $O\big( \mathit{n\_epochs} * \mathit{n\_batches} * \mathit{b} * (Dh_1+h_1h_2+\cdots+h_H) \big)$ for its training and $O\big( I(Dh_1+h_1h_2+\cdots+h_H) \big)$ for its testing, where $I$ is the data size of the test set.


\subsection{Interpretability of Anomaly Scores}

At the testing stage, like other anomaly detection methods, DevNet uses the optimized $\phi$ to produce an anomaly score for every test object and returns an anomaly ranking of the data objects based on the anomaly scores, in which the top-ranked objects are anomalies. However, the anomaly scores returned by most anomaly detectors are often not easily interpretable \cite{kriegel2011interpreting}. As a result, given a data object's anomaly score, it is not clear what is the probability of this object being an anomaly, and it is also difficult to determine a specific threshold to select the appropriate top-ranked objects. Therefore, if users need more than an anomaly ranking in practice, some types of separate anomaly score unification methods \cite{kriegel2011interpreting} are required for those methods to transform their scores into more interpretable ones. However, the anomaly scoring and the score unification are two independent modules in such cases, which may lead to untrustworthy explanation of the scores. By contrast, DevNet directly yields highly interpretable anomaly scores.

\begin{proposition}
Let $\mathbf{x} \in \mathcal{X}$ and $z_{p}$ be the quantile function of $\mathcal{N}(\mu, \sigma^{2})$, then $\phi(\mathbf{x})$ lies outside the interval $\mu \pm z_{p}\sigma $ with a probability $2(1-p)$.
\end{proposition}

This proposition of DevNet is due to the Gaussian prior and Z-Score-based deviation loss. The probability $2(1-p)$ offers a straightforward explanation to the anomalousness of any given score $\phi(\mathbf{x})$. Particularly, we have the probability $(1-p)$ when only focusing on the upper tail $\mu + z_{p}\sigma $, e.g., by applying $p=0.95$, we have $z_{0.95}=1.96$, which states that having anomaly scores over 1.96 (as $\mu=0$ and $ \sigma=1 $ are used in DevNet) indicates the object only has a probability of 0.05 generated from the same mechanism as the normal data objects. Users can also easily choose a threshold to determine anomalies with a desired confidence level, e.g., given the anomaly score distribution shown in Figure \ref{fig:example}(b), it is easy to use $z_{0.95}$ to identify the anomalies with a 95\% confidence level.



\section{Experiments}

\subsection{Data Sets}

As shown in Table \ref{tab:rocpr}, nine publicly available real-world data sets are used, which are from diverse critical domains, e.g., intrusion detection, fraud detection, malicious URL detection, and disease detection. Five data sets contain real anomalies, i.e., exceptionally exciting projects in \textit{donors}, fraudulent credit card transactions in \textit{fraud}, backdoor network attacks in \textit{backdoor}, malicious URLs in \textit{URL}, and hypothyroid patients in \textit{thyroid}. The other four data sets contain semantically real anomalies, i.e., they are rare or very different from the majority of data objects. The detailed information of accessing and preprocessing the data sets can be found in Appendix \ref{sec:preprocessing}.

\begin{table*}[htbp]
  \centering
  \caption{AUC-ROC and AUC-PR Performance (with $\pm$ standard deviation) of DevNet and Four Competing Methods. \#obj. is the overall data size, $D$ is the dimensionality size, $f_1$ and $f_2$ denote the percentage that the labeled anomalies respectively comprise in the training data and the total anomalies. $D$ in \textit{URL} and \textit{news20}, i.e., `3M' and `1M', are short for 3,231,961 and 1,355,191, respectively. The best performance is boldfaced. }
        
  \scalebox{0.78}{
    \begin{tabular}{||p{0.95cm}p{0.7cm}p{0.2cm}p{0.45cm}p{0.55cm}||p{1.5cm}p{1.4cm}p{1.4cm}p{1.4cm}p{1.4cm}||p{1.5cm}p{1.4cm}p{1.4cm}p{1.4cm}c||}
    \hline
    \multicolumn{5}{||c||}{\textbf{Data Characteristic}} & \multicolumn{5}{|c||}{\textbf{AUC-ROC Performance}}               & \multicolumn{5}{c||}{\textbf{AUC-PR Performance}} \\
    \hline
     \textbf{Data}      & \#obj.      & $D$     & \centering$f_1$ &  \centering $f_2$ & \centering \textbf{DevNet} & \centering \textbf{REPEN} & \centering \textbf{DSVDD} & \centering \textbf{FSNet} & \centering \textbf{iForest} & \centering \textbf{DevNet} & \centering \textbf{REPEN} & \centering \textbf{DSVDD} & \centering \textbf{FSNet} &  \textbf{iForest} \\\hline
    donors & 619,326 & 10 & 0.01\% & 0.08\% & \textbf{1.000}$\pm$0.000 & 0.975$\pm$0.005 & 0.995$\pm$0.005 & 0.997$\pm$0.002 & 0.874$\pm$0.015 & \textbf{1.000}$\pm$0.000 & 0.508$\pm$0.048 & 0.846$\pm$0.114 & 0.994$\pm$0.002 & 0.221$\pm$0.025 \\ 
    census & 299,285 & 500 & 0.01\% & 0.16\% & 0.828$\pm$0.008 & 0.794$\pm$0.005 & \textbf{0.835}$\pm$0.014 & 0.732$\pm$0.020 & 0.624$\pm$0.020 & \textbf{0.321}$\pm$0.004 & 0.164$\pm$0.003 & 0.291$\pm$0.008 & 0.193$\pm$0.019 & 0.076$\pm$0.004 \\ 
    fraud & 284,807 & 29 & 0.01\% & 6.10\% & \textbf{0.980}$\pm$0.001 & 0.972$\pm$0.003 & 0.977$\pm$0.001 & 0.734$\pm$0.046 & 0.953$\pm$0.002 & \textbf{0.690}$\pm$0.002 & 0.674$\pm$0.004 & 0.688$\pm$0.004 & 0.043$\pm$0.021 & 0.254$\pm$0.043 \\ 
    celeba & 202,599 & 39 & 0.02\% & 0.66\% & \textbf{0.951}$\pm$0.001 & 0.894$\pm$0.005 & 0.944$\pm$0.003 & 0.808$\pm$0.027 & 0.698$\pm$0.020 & \textbf{0.279}$\pm$0.009 & 0.161$\pm$0.006 & 0.261$\pm$0.008 & 0.085$\pm$0.012 & 0.065$\pm$0.006 \\ 
    backdoor & 95,329 & 196 & 0.04\% & 1.29\% & \textbf{0.969}$\pm$0.004 & 0.878$\pm$0.007 & 0.952$\pm$0.018 & 0.928$\pm$0.019 & 0.752$\pm$0.021 & \textbf{0.883}$\pm$0.008 & 0.116$\pm$0.003 & 0.856$\pm$0.016 & 0.573$\pm$0.167 & 0.051$\pm$0.005 \\ 
    URL & 89,063 & 3M & 0.04\% & 1.69\% & \textbf{0.977}$\pm$0.004 & 0.842$\pm$0.006 & 0.908$\pm$0.027 & 0.786$\pm$0.047 & 0.720$\pm$0.032 & \textbf{0.681}$\pm$0.022 & 0.103$\pm$0.003 & 0.475$\pm$0.040 & 0.149$\pm$0.076 & 0.066$\pm$0.012 \\ 
    campaign & 41,188 & 62 & 0.10\% & 0.65\% & \textbf{0.807}$\pm$0.006 & 0.723$\pm$0.006 & 0.748$\pm$0.019 & 0.623$\pm$0.024 & 0.731$\pm$0.015 & \textbf{0.381}$\pm$0.008 & 0.330$\pm$0.009 & 0.349$\pm$0.023 & 0.193$\pm$0.012 & 0.328$\pm$0.022 \\ 
    news20 & 10,523 & 1M & 0.37\% & 5.70\% & \textbf{0.950}$\pm$0.007 & 0.885$\pm$0.003 & 0.887$\pm$0.000 & 0.578$\pm$0.050 & 0.328$\pm$0.016 & \textbf{0.653}$\pm$0.009 & 0.222$\pm$0.004 & 0.253$\pm$0.001 & 0.082$\pm$0.010 & 0.035$\pm$0.002 \\ 
    thyroid & 7,200 & 21 & 0.55\% & 5.62\% & \textbf{0.783}$\pm$0.003 & 0.580$\pm$0.016 & 0.749$\pm$0.011 & 0.564$\pm$0.017 & 0.688$\pm$0.020 & \textbf{0.274}$\pm$0.011 & 0.093$\pm$0.005 & 0.241$\pm$0.009 & 0.116$\pm$0.014 & 0.166$\pm$0.017 \\ \hline
          \multicolumn{5}{||r||}{ Average} & \textbf{0.916}$\pm$0.004 &	0.838$\pm$0.006 &	0.888$\pm$0.011 &	0.750$\pm$0.028	& 0.708$\pm$0.018 &	\textbf{0.574}$\pm$0.008	& 0.263$\pm$0.010 &	0.473$\pm$0.025 &	0.270$\pm$0.037 & 0.140$\pm$0.015 \\
           \multicolumn{5}{||r||}{P-value} &  \centering -    & \centering 0.004 & \centering 0.023 & \centering 0.004 & \centering 0.004 &  \centering -    & \centering 0.004 & \centering 0.004 & \centering 0.004 & 0.004 \\

    \hline
    \end{tabular}%
    }
  \label{tab:rocpr}%
\end{table*}%

\subsection{Competing Methods}

DevNet is compared with four methods, including REPEN \cite{pang2018repen}, \textit{adaptive} Deep SVDD (DSVDD) \cite{ruff2018deepsvdd}, prototypical networks (denoted as FSNet) \cite{snell2017protonet}, and iForest \cite{liu2012iforest}. These four methods are chosen because they are the state-of-the-art in the relevant areas, i.e., REPEN in deep anomaly detection with limited labeled data, DSVDD in feature learning for anomaly detection, FSNet in few-shot classification, and iForest in unsupervised anomaly detection. 

The original DSVDD is designed to minimize the distance between a fixed one-class center vector $\mathbf{c}$ and the training data in the projected space, in which the labeled anomalies cannot be used. To have a fair comparison to DevNet, we modified DSVDD to fully leverage the labeled anomalies based on \cite{tax2004svdd}, by adding an additional term into its objective function to guarantee a large margin between normal objects and anomalies in the new space while minimizing the $\mathbf{c}$-based hypersphere's volume. This adaption significantly enhances the original DSVDD. 

All methods are implemented in Python: DevNet, DSVDD and FSNet are implemented using Keras \cite{chollet2015keras}, REPEN is taken from its authors and is also built upon Keras, and iForest is taken from the scikit-learn package.

\subsection{Parameter Settings}\label{sec:parameter}
Since our experiments focus on unordered multidimensional data, multilayer perceptron (MLP) network architectures are used. Specifically, we tested two architectures for all neural methods. Motivated by the success of REPEN \cite{pang2018repen}, our first network uses the same architecture as REPEN, i.e., one hidden layer with 20 neural units. The second architecture consists of three hidden layers to learn more intricate data interactions, with 1,000 units in the first hidden layer, followed by 250 and 20 units in the second and third hidden layers, respectively. The ReLu function $g(z) = \mathit{max}(0, z)$ is used because of its efficient computation and gradient propagation, and an $\ell_2$-norm regularizer is applied to every hidden layer to avoid overfitting.

All DevNet, REPEN, DSVDD and FSNet were tested using these two architectures on all the data sets, and we found all of them performed best with the one hidden layer structure. This may be due to the limit of the available labeled data. Due to the page limitation, we report the results based on the architecture with one hidden layer by default. We show the results of DevNet using the three hidden layers in our ablation study in Section \ref{exp:ablation}.



In training, DevNet, DSVDD and FSNet use the Root Mean Square propagation (RMSprop) optimizer \cite{hinton2012rmsprop} to perform gradient descents, and they are trained using 50 epochs, with 20 min-batches in each epoch. These settings enable the three deep detectors to achieve stable performance across the data sets. iForest is a non-neural ensemble method. It is used with the recommended settings, i.e., subsampling size set to 256 and ensemble size set to 100  \cite{liu2012iforest}. iForest cannot work in data with millions of features, so we use the sparse random projection \cite{li2006srp} to map \textit{URL} and \textit{news20} into a 1,000-dimensional space before applying iForest, which obtains better performance than other projection options.

\subsection{Performance Evaluation Methods}
We use two popular and complementary performance metrics, the Area Under Receiver Operating Characteristic Curve (AUC-ROC) and the Area Under Precision-Recall Curve (AUC-PR), to have a comprehensive evaluation of anomaly detectors. AUC-ROC summarizes the ROC curve of true positives against false positives, while AUC-PR is a summarization of the curve of precision against recall. Specifically, an AUC-ROC value of one indicates the best performance, while a value close to 0.5 indicates a random ranking of the objects. AUC-ROC is widely used due to its good interpretability. 

However, AUC-PR is more suitable than AUC-ROC in many anomaly detection applications which require excellent performance on the positive class and do not care much of the performance on the negative class. This is because AUC-ROC is affected by the performance on both anomaly and normal classes and the performance on the normal class can bias AUC-ROC due to the class-imbalance nature of anomaly detection data. By contrast, AUC-PR evaluates how many positive predictions are correct (precision), and how many of the positive predictions that are truly positive compose the positive class (recall). We use a widely-used method, known as average precision in \cite{schutze2008introduction}, to calculate AUC-PR. Large AUC-PR indicates better performance, but it is often very challenging to achieve large AUC-PR due to the skewed and heterogeneous distributions of anomalies.

The reported AUC-ROC, AUC-PR, and runtimes are averaged results over 10 independent runs. The paired \textit{Wilcoxon} signed rank test \cite{woolson2007wilcoxon} is used to examine the significance of the performance of DevNet against its competing methods. All runtimes are calculated at a node in a 2.4GHz Phoenix cluster with 64GB dedicated memory using 8 cores and 1 Tesla K80 GPU accelerator. 

\subsection{Effectiveness in Real-world Data Sets}\label{exp:effectiveness}

\subsubsection{Experiment Settings}

This section examines the performance of DevNet on common real-life application scenarios where there are a large number of unlabeled data objects with a very small set of labeled anomalies. To replicate such scenarios, the anomalies and normal objects in each data set are first splitted into two subsets, with 80\% data as training data and the other 20\% data as test set. To have controlled experiments w.r.t. anomaly contamination, we then randomly add/remove the anomalies in each training data set such that the anomalies account for 2\% of the training data, i.e., 2\% anomaly contamination (other contamination levels are further examined in Section \ref{exp:contamination}). The resulted data forms the unlabeled training data set $\mathcal{U}$. We further randomly sample 30 anomalies from the anomaly class as the prior knowledge of the anomalies of interest, i.e., the labeled anomaly set $\mathcal{K}$, which accounts for only 0.005\%-1\% of all training data objects and 0.08\%-6\% of the anomaly class (see $f_1$ and $f_2$ in Table \ref{tab:rocpr} for detail). Since only the class label of $\mathcal{K}$ is used during training, the task is equivalent to unsupervised anomaly detection with a few additional labeled anomalies available as prior knowledge. We will investigate the detection performance w.r.t. different amount of the prior knowledge in Section \ref{exp:labeledanomalies}. 

\subsubsection{Findings - The direct optimization of anomaly scores enables DevNet to achieve significant improvement over other deep methods}
The AUC-ROC and AUC-PR performance of DevNet and four competing methods are shown in Table \ref{tab:rocpr}. DevNet performs best on eight and nine data sets in the respective AUC-ROC and AUC-PR performance, and it performs comparably well to the best performer on \textit{census} in AUC-ROC where it ranks in second. In terms of AUC-ROC, DevNet obtains substantially better average improvement than REPEN (9\%), DSVDD (3\%), FSNet (22\%) and iForest (29\%) and the improvement is statistically significant at the 95\% or 99\% confidence interval; in terms of AUC-PR, the improvement DevNet achieves is much more substantial than REPEN (118\%), DSVDD (21\%), FSNet (113\%) and iForest (309\%), which is all statistically significant at the 99\% confidence interval. These results are due to the reason that DevNet efficiently leverages the limited available anomalies to well optimize the anomaly scores, resulting in high-quality anomaly rankings, i.e., substantially high precision and recall of detecting anomalies; while the competing methods have an indirect learning of anomaly scores, resulting in weak capability of discriminating some intricate anomalies from normal objects and thus high false positives and low recall rates.

\subsection{Data Efficiency}\label{exp:labeledanomalies}

\subsubsection{Experiment Settings}
This section examines the data efficiency of the deep methods, which is a critical factor as it is very difficult to obtain labeled anomalies in most anomaly detection applications. The number of available labeled anomalies varies from 5 to 120, with the anomaly contamination level fixed to be 2\%. iForest is used as the baseline, which is an unsupervised method and thus insensitive to the amount of the labeled data. We aim to answer the following two key questions:
\begin{itemize}
    \item How data-efficient are the DevNet and other deep methods?
    \item How much improvement can the deep methods gain from the labeled anomalies compared to the unsupervised iForest?
\end{itemize}

\subsubsection{Findings - DevNet is the most data-efficient method; and the improvement due to the limited labeled anomalies is very substantial}
\begin{figure}[h!]
  \centering
    \includegraphics[width=0.48\textwidth]{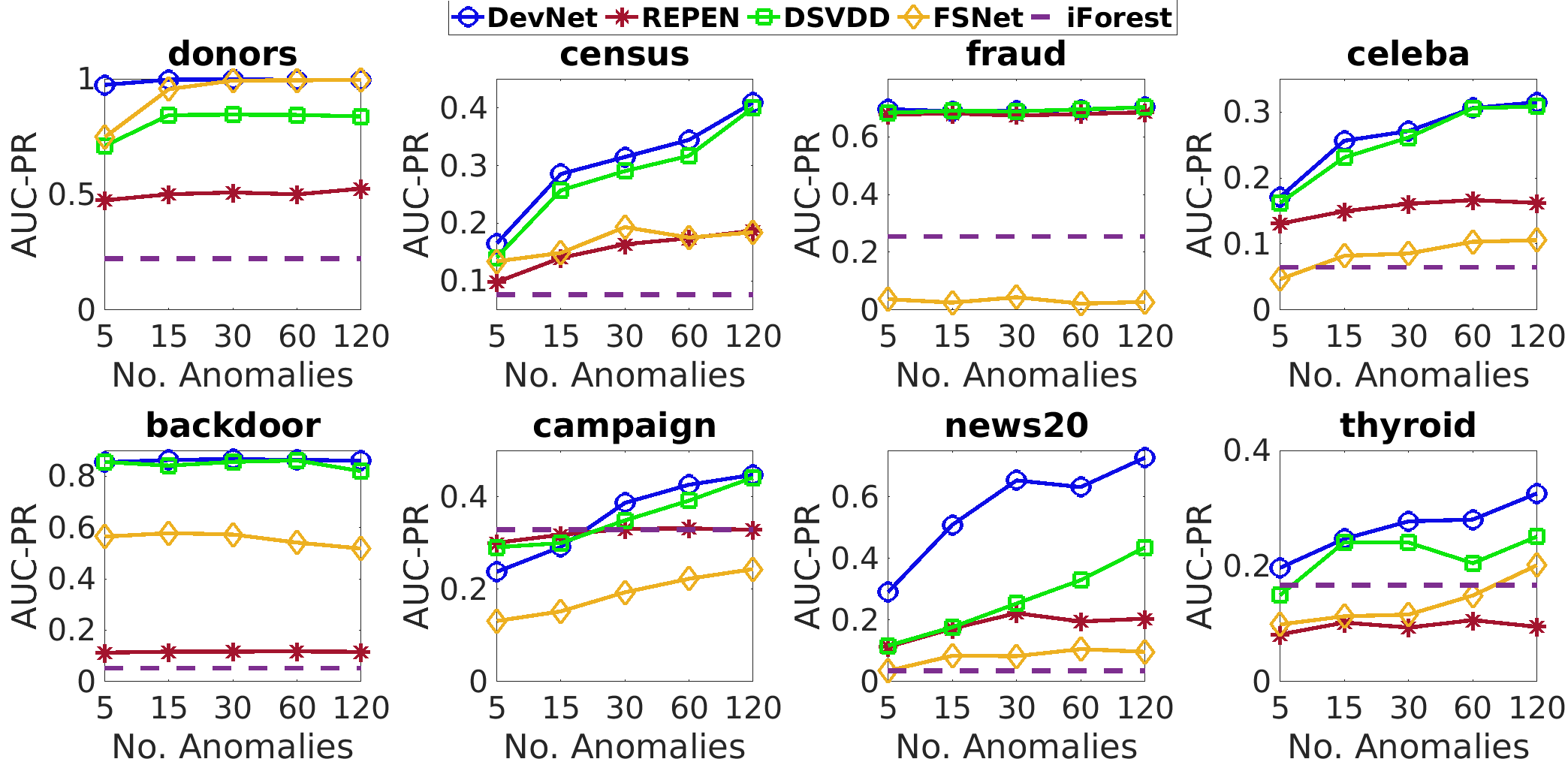}
  \caption{AUC-PR w.r.t. No. Labeled Anomalies. The results on \textit{URL} are omitted due to prohibitively expensive computation.}
  \label{fig:knownouliers}
\end{figure}

Figure \ref{fig:knownouliers} shows the AUC-PR results w.r.t. different number of labeled anomalies available. Similar results can also be observed in AUC-ROC. The performance of these four deep methods generally increases with increasing number of labeled anomalies, since more labeled data generally helps train the model better. However, the AUC-PR of some competing deep detectors drops with more labeled data in some cases, e.g., FSNet in \textit{census} and \textit{backdoor}, REPEN in \textit{celeba} and \textit{news20}, DSVDD in \textit{backdoor} and \textit{thyroid}. This may be due to the scattered and dissimilar distributions of anomalies, because when the added labeled anomalies have very different anomalous behaviors and carry information conflicting to the other labeled anomalies for the optimization, they may then downgrade the detection performance. Compared to the counterparts, DevNet is more stable in such cases.

DevNet is the most data-efficient method, which obtains the best average performance w.r.t. different number of labeled anomalies and achieves the fastest increase rate of AUC-PR against the number of labeled anomalies. Impressively, DevNet needs 75\%-88\% less labeled data to achieve comparably better performance to the best competing method in several cases, e.g., DevNet requires 83\% less labeled data to achieve comparably good performance to the best contender FSNet on \textit{donors}, and outperforms the best contender DSVDD on \textit{news20} and \textit{thyroid} using respective 88\% and 75\% less labeled data. The DevNet's superiority is due to its end-to-end differentiable learning of the anomaly scores, because it allows DevNet to directly optimize the anomaly scores with the limited labeled data, which can leverage the data much more efficiently than the counterpart two-step approach.

Compared to the unsupervised method iForest, even when only a very few labeled anomalies (e.g., 5 or 15) are used, the improvement of the prior knowledge-driven deep methods, especially DevNet and DSVDD, is very substantial on most data sets, such as \textit{donors}, \textit{census}, \textit{fraud}, \textit{celeba}, \textit{backdoor}, \textit{news20} and \textit{thyroid}; for example, the average improvement of DevNet and DSVDD using 5 labels over iForest is more than 400\%. In the case of \textit{campaign} that may have very intricate distributions of anomalies, the deep methods need slightly more labeled data to achieve the similarly large improvement.

\subsection{Robustness w.r.t. Anomaly Contamination}\label{exp:contamination}

\subsubsection{Experiment Settings}
Recall that we use a simple training strategy to train DevNet and the other deep methods, i.e., all unlabeled training data objects in $\mathcal{U}$ are used as normal data objects and we sample negative data objects from this set of objects to comprise a half of data objects in each mini-batch (see Step 4 in Algorithm \ref{alg:devnet}). This section investigates the robustness of DevNet w.r.t. different anomaly contamination levels in the unlabeled training data. We vary the contamination level from 0\% up to 20\%, with the number of available labeled anomalies fixed to be 30. We aim to examine the following two key questions: 
\begin{itemize}
    \item How robust are the deep anomaly detectors?
    \item Can the deep methods still substantially beat the unsupervised method iForest when the contamination level is high? 
\end{itemize}

\subsubsection{Findings - DevNet is consistently more robust than the other deep methods; and the substantially better improvement of DevNet over iForest persists even when a very large anomaly contamination is presented in the unlabeled training data}

\begin{figure}[h!]
  \centering
    \includegraphics[width=0.48\textwidth]{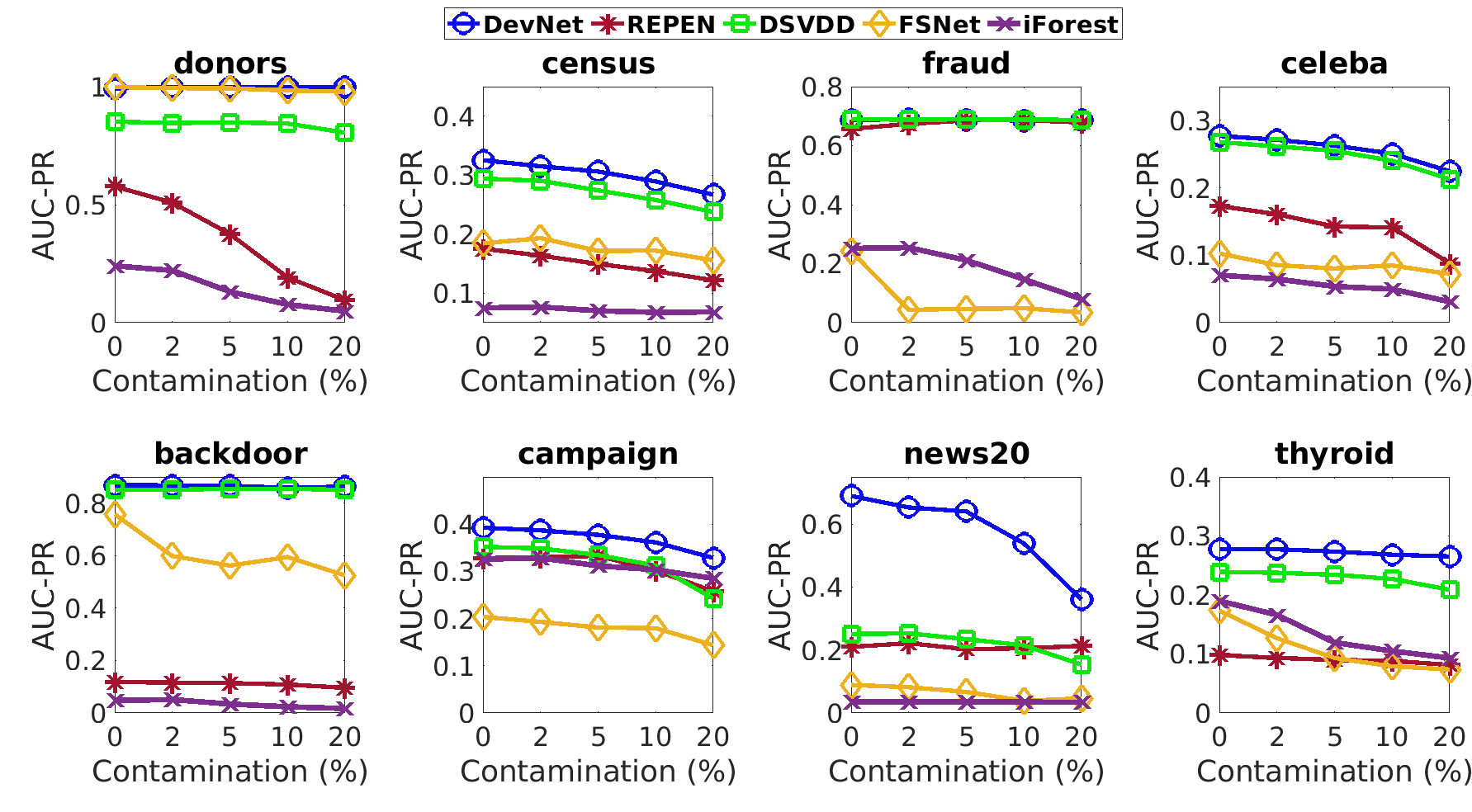}
  \caption{AUC-PR w.r.t. Different Contamination Rates. The results on \textit{URL} are omitted due to prohibitively expensive computation.}
  \label{fig:contrate}
\end{figure}

The AUC-PR results w.r.t. different anomaly contamination levels are presented in Figure \ref{fig:contrate}. Similar results can also be observed in AUC-ROC. The performance of all deep anomaly detectors decreases with increasing contamination levels. This is because the probability of falsely sampling anomalies from the unlabeled data as normal objects gets larger in the mini-batch construction, which can mislead the stochastic gradient descent-based optimization and downgrade the detection accuracy. Nevertheless, it is clear that DevNet performs consistently better and achieves remarkably better average AUC-PR performance than REPEN (200\%), DSVDD (28\%) and FSNet (336\%) over the different contamination levels. This demonstrates a strong capability of DevNet in tapping the limited prior knowledge to well optimize the anomaly scores in challenging noisy environments.

Compared to iForest, the four deep methods obtain substantially better average AUC-PR improvement across the eight data sets, e.g., DevNet and DSVDD have respectively more than 800\% and 600\% average improvement. This is because although the large anomaly contamination in the unlabeled data presents many noises to the deep model training, the small set of labeled anomalies empowers the deep methods and help them to largely defy the noises. By contrast, the unsupervised method iForest does not have any prior knowledge of anomalies and thus returns many noisy or uninteresting objects as anomalies, leading to very large false positives; also, its performance still decreases with increasing anomaly contamination rate, because the unsupervised methods like iForest typically assume that anomalies are rare in the unlabeled data and thus they perform less effectively when the increasing anomaly contamination violates the assumption.

\subsection{Ablation Study}\label{exp:ablation}

\subsubsection{Experiment Settings}
We examine the importance of the key components of DevNet by comparing DevNet to its three variants. Recall that the default DevNet (denoted as Def) has one hidden layer with 20 ReLu units and a linear unit in the output layer. 
\begin{itemize}
    \item The first variant is DevNet-Rep, which removes the output layer of Def and uses our deviation loss to learn the representations only. In this case, the reference in the loss function is a 20-dimensional vector rather than a scalar.
    \item The second variant is DevNet-Linear, which removes the non-linear learning hidden layer of Def, making it equivalent to learning a direct linear mapping from the original data space to the anomaly score space.
    \item The third variant is DevNet-3HL, in which three hidden layers with respective 1000, 250 and 20 ReLu units are used.
\end{itemize}

\subsubsection{Findings - The end-to-end learning of anomaly scores, deviation loss, and learning of non-linear features all have some major contributions to the superior performance of DevNet}

Table \ref{tab:ablation} shows the performance of DevNet and its three variants. The end-to-end learning of anomaly scores enables Def to obtain more accurate and stable performance than Rep that focuses on feature learning. Def performs less effectively than Rep in \textit{census}. This may be due to that some normal objects and anomalies in \textit{census} are quite similar, which can mislead the score learning in Def more severely than the representation learning in Rep.

\begin{table}[htbp]
\caption{AUC-ROC and AUC-PR Results of DevNet and Its Variants.}
\scalebox{0.78}{
    \begin{tabular}{||l||cccc||cccc||}
    \hline
     & \multicolumn{4}{|c||}{\textbf{AUC-ROC Performance}}               & \multicolumn{4}{c||}{\textbf{AUC-PR Performance}} \\
    \hline
    Data & Def & Rep & Linear & 3HL  & Def & Rep & Linear & 3HL  \\ \hline
    donors & \textbf{1.000} & 0.999 & 0.978 & \textbf{1.000} & \textbf{1.000} & 0.976 & 0.827 & \textbf{1.000} \\
    census & 0.828 & \textbf{0.858} & 0.832 & 0.686 & 0.321 & \textbf{0.338} & 0.297 & 0.241 \\
    fraud & \textbf{0.980} & 0.975 & 0.937 & 0.926 & 0.690 & 0.684 & 0.659 & \textbf{0.701} \\
    celeba & \textbf{0.951} & 0.949 & 0.949 & 0.877 & 0.279 & \textbf{0.283} & 0.281 & 0.239 \\
    backdoor & \textbf{0.969} & 0.913 & 0.928 & 0.968 & \textbf{0.883} & 0.846 & 0.555 & 0.843 \\
    URL & \textbf{0.977} & 0.954 & 0.872 & 0.941 & 0.681 & \textbf{0.687} & 0.347 & 0.595 \\
    campaign & \textbf{0.807} & 0.759 & 0.757 & 0.679 & \textbf{0.381} & 0.371 & 0.357 & 0.259 \\
    news20 & 0.950 & \textbf{0.953} & 0.819 & 0.817 & \textbf{0.653} & 0.552 & 0.447 & 0.421 \\
    thyroid & 0.783 & 0.729 & 0.717 & \textbf{0.787} & 0.274 & 0.216 & 0.205 & \textbf{0.383} \\ \hline
    Average & \textbf{0.916} & 0.899 & 0.865 & 0.853 & \textbf{0.574} & 0.550 & 0.442 & 0.520 \\ \hline
    P-value &  - & 0.129 & 0.012 & 0.023 &  - & 0.106 & 0.008 & 0.133 \\ \hline

\end{tabular}
}
\label{tab:ablation}
\end{table}

Note that Rep and DSVDD actually share a similar objective, but Rep uses the deviation loss while DSVDD uses the SVDD-based loss. Compared to DSVDD in Table \ref{tab:rocpr}, Rep performs slightly better in AUC-ROC (1\% improvement) and substantially better in AUC-PR (16\% improvement). This indicates that our deviation loss offers a much better capability in capturing different anomalous behaviors.

Compared to Linear, Def obtains significantly better average AUC-ROC (6\%) and AUC-PR (30\%) improvement, indicating a significant role of the intermediate non-linear feature learning before the learning of the anomaly scores. However, as illustrated by the substantial average improvement of Def over 3HL, deepening the hidden layers from one layer to three layers is not always beneficial, because we have only a few labeled anomalies, which are often not sufficient to well train a deeper model.

\subsection{Scalability Test}

\subsubsection{Experiment Settings} 
We examine the scalability w.r.t. data size by generating four synthetic 1,000-dimensional data sets with varying data sizes. Similarly, the scaleup test w.r.t. dimension uses a fixed data size (i.e., 5,000) and varying dimensions. Each detector is trained and tested in a data set of the same size. The runtime below includes both training and testing execution time.

\subsubsection{Findings - DevNet has a linear time complexity w.r.t. both data size and dimension}

\begin{figure}[h!]
  \centering
    \includegraphics[width=0.45\textwidth]{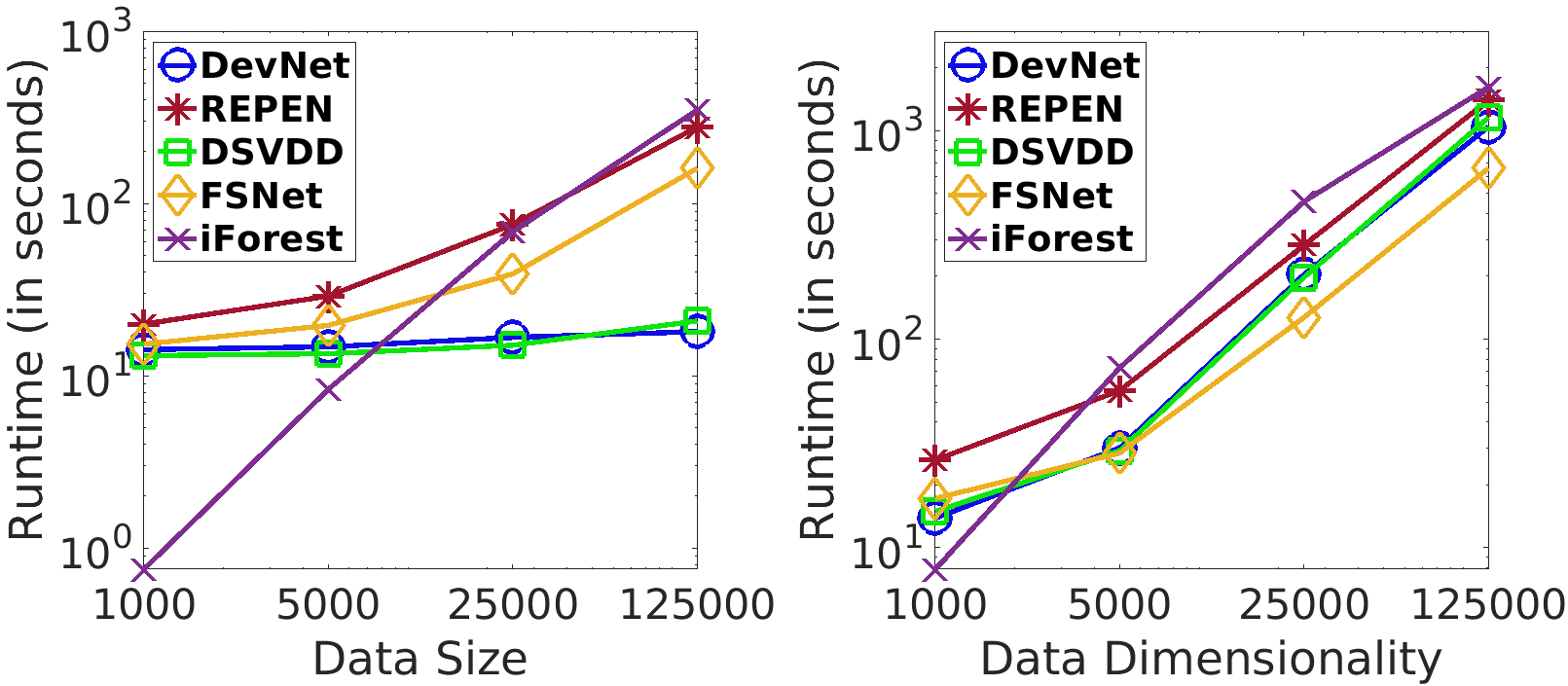}
  \caption{Scalability Test w.r.t. Data Size and Dimensionality.}
  \label{fig:scaleup}
\end{figure}

The scaleup test results are presented in Figure \ref{fig:scaleup}. These results show that the overall runtime of DevNet increases linearly w.r.t. both data size and dimension, which justifies the complexity analysis w.r.t. multilayer perceptron networks in Section \ref{sec:algo}. Particularly, although REPEN, FSNet and iForest also have linear time complexity, DevNet runs considerably faster than them by a factor of 10 to 20 on the large data sets. This is because the loss function in DevNet is very computationally efficient, whereas REPEN and FSNet involves extensive distance computation in both training and testing, and iForest needs much time on constructing isolation trees. On the high-dimensional data, DevNet runs comparably fast to REPEN and DSVDD but slightly slower than FSNet. This may be due to the fact that the computation in the bottom layers that project original very high-dimensional data into low-dimensional space dominates the overall runtime, as it is much more costly than the top layer that calculates the loss. As a result, FSNet, which uses a much smaller mini-batch size, requires less time to process each batch data and obtains a better computation efficiency than other methods like DevNet and DSVDD. iForest requires considerable time to perform random data space partition when the dimension is large, leading to the most costly method here.

\section{Conclusions}

This paper introduces a novel framework and its instantiation DevNet for leveraging a few labeled anomalies with a prior to fulfill an end-to-end differentiable learning of anomaly scores. By a direct optimization of anomaly scores, DevNet can be trained much more data-efficiently, and performs significantly better in terms of both AUC-ROC and AUC-PR, compared to the two-step deep anomaly detectors that focus on optimizing feature representations. We also find empirically that deep anomaly detectors can be well trained by randomly sampling negative examples from the anomaly-contaminated unlabeled data and positive examples from the small labeled anomaly set. Even when the anomaly contamination level is high, the deep detectors, especially DevNet, can still perform very well and achieve significant improvement over the state-of-the-art unsupervised anomaly detectors. This may provide a new perspective for optimizing anomaly detection methods.

We are testing DevNet on image and sequence data using convolutional/recurrent network architectures, and plan to extend DevNet by a hybrid of data-driven and prior-driven reference score generation approach for extremely challenging real-world applications where only one or two labeled anomalies are available.

\section*{Acknowledgements}
This work is partially supported by the ARC Discovery Project DP180103023.

\bibliographystyle{ACM-Reference-Format}
\balance
\bibliography{references}

\appendix
\section{supplementary Material for Reproducibility}

\subsection{Data Accessing and Preprocessing}\label{sec:preprocessing}
The \textit{donors} data is taken from KDD Cup 2014 for predicting excitement of projects proposed by K-12 school teachers, in which exceptionally exciting projects are used as anomalies (5.92\% data). The \textit{census} data is extracted from the US census bureau database, in which we aim to detect the rare high-income person (i.e., the person who earns over 50K dollars a year), which is about 6\% of the data. The \textit{fraud} data is for fraudulent credit card transaction detection, in which the fraudulent transactions are used as anomalies. The \textit{celeba} data is a large-scale image data set which contains more than 200K celebrity images, each with 40 attribute annotations. We use the \textit{bald} attribute as our detection target, in which the scarce bald celebrities, less than 3\% celebrities, are treated as anomalies, and the other 39 attributes form the learning feature space. The \textit{backdoor} data is a backdoor attack detection data set with the attacks as anomalies against the `normal' class, which is extracted from the UNSW-NB 15 data set \cite{moustafa2015nb15}. The \textit{URL} data is for malicious URL detection, which consists of 120-day collection of malicious and benign URLs \cite{ma2009url}. Following \cite{pang2018repen}, the first-week subset of this collection is used and the malicious URLs are used as anomalies. The \textit{campaign} data is a data set of direct bank marketing campaigns via phone calls, in which the rarely successful campaigning records, accounting for about 10\% records, are used as anomalies. The \textit{news20} data is a balanced text classification data set. Following the literature \cite{keller2012hics,pang2018repen}, \textit{news20} is converted to anomaly detection data with 5\% anomalies by downsampling the small class. The \textit{thyroid} data is a disease detection data set, in which the anomalies are the patients diagnosed with hypothyroid. All these data sets can be publicly accessed via the links provided in Table \ref{tab:link}.

\begin{table}[htbp]
\caption{Links for Accessing the Data Sets}
\scalebox{0.78}{
    \begin{tabular}{|l|p{8.5cm}|}
    \hline
    \textbf{Data} & \textbf{Link}\\\hline
    donors&https://www.kaggle.com/c/kdd-cup-2014-predicting-excitement-at-donors-choose \\ 
    census&https://archive.ics.uci.edu/ml/datasets/Census-Income+(KDD) \\
    fraud&https://www.kaggle.com/mlg-ulb/creditcardfraud \\ 
    celeba&http://mmlab.ie.cuhk.edu.hk/projects/CelebA.html \\ 
    backdoor&https://www.unsw.adfa.edu.au/unsw-canberra-cyber/cybersecurity/ADFA-NB15-Datasets/ \\ 
    URL&http://www.sysnet.ucsd.edu/projects/url/ \\ 
    campaign&https://archive.ics.uci.edu/ml/datasets/bank+marketing\\
    news20&https://www.csie.ntu.edu.tw/$\sim$cjlin/libsvmtools/datasets/binary/ \\ 
    thyroid&http://archive.ics.uci.edu/ml/datasets/thyroid+disease \\ \hline
    \end{tabular}
}
\label{tab:link}
\end{table}

For these data sets, missing values are replaced with the mean value in the corresponding feature, and categorical features are encoded by one-hot encoding. 

\subsection{Algorithm Implementation}
This section provides the detailed information of our implementation of algorithms. Relevant key information is also presented in Section \ref{sec:parameter}.

\subsubsection{Implementation of Competing Methods}
We use the implementation of iForest available at the scikit-learn Python package. REPEN is directly taken from the authors. Its codes are publicly accessible at https://sites.google.com/site/gspangsite/sourcecode. We implement and further enhance DSVDD by adding an additional margin term into the one-class SVDD objective to enforce a margin between the center $\mathbf{c}$ and the labeled anomalies in the new representation space. Similar to DevNet, the contrastive loss \cite{hadsell2006contrastloss} is used in DSVDD to fulfill this margin-based optimization. The anomaly score is defined as the distance to the one-class center $\mathbf{c}$, which is exactly the same as in its original paper. Due to the incorporating of the few labeled anomalies, the modified DSVDD substantially improves the original DSVDD by more than 30\% detection accuracy. For FSNet, since we do not have the finer-grained class information in the training data, we cannot construct the training episodes in the same way as in \cite{snell2017protonet}. Instead we randomly sample the same number of data objects from the unlabeled training data and from the limited labeled anomalies to form the desired episodes for training FSNet. The anomaly score is then calculated as a softmax over distances to the respective normal and anomaly prototypes.

\subsubsection{Optimization Settings} In optimizing the deep anomaly detection methods, the default settings of the layers or optimizer in Keras are used, and they are as described in Section \ref{sec:parameter} otherwise. Particularly, for the hidden layer, we use the dense layer with an uniform Glorot weight initialization and an $\ell_2$-norm weight decay regularizer (as recommended in Keras, the hyperparameter setting $\lambda=0.01$ is used in the regularizer). No constraints are applied to the kernels or biases. The activation function is the default ReLu function. The Root Mean Square propagation (RMSprop) optimizer is used with the recommended settings in Keras, i.e., $\mathit{lr}=0.001$, $\rho=0.9$, $\epsilon=\mathit{None}$, and $\mathit{decay}=0.0$. The mini-batch size is probed using a set of commonly used options, $\{8, 16, 32, 64, 128, 256, 512\}$. The best fits, 512 in DevNet and DSVDD, and 256 in FSNet, are used by default. Since REPEN was designed for a similar problem scenario as DevNet, it is used with the recommended optimization settings as in \cite{pang2018repen}.

\subsubsection{Packages Used in Our Implementation}
The relevant packages and their versions used in our algorithm implementation are listed as follows:
\begin{itemize}
    \item python==3.6.6
    \item keras==2.2.4
    \item keras-applications==1.0.6
    \item keras-preprocessing==1.0.5
    \item tensorflow-gpu==1.10.0
    \item scikit-learn==0.20.0
    \item numpy==1.14.5
    \item pandas==0.23.4
    \item scipy==1.1.0
    \item tensorboard==1.10.0
\end{itemize}

\end{document}